\documentclass{article}
\usepackage{spconf,amsmath,epsfig}
\usepackage{times}
\usepackage{graphicx}
\usepackage{amssymb}
\usepackage{subfigure}
\usepackage{tabularx}
\usepackage{float}
\usepackage{caption}
\usepackage{xcolor}

\let\OLDthebibliography\thebibliography
\renewcommand\thebibliography[1]{
  \OLDthebibliography{#1}
  \setlength{\parskip}{0pt}
  \setlength{\itemsep}{0pt plus 0.3ex}
}

\begin{document}\sloppy

\def\x{{\mathbf x}}
\def\L{{\cal L}}

\title{Towards Learning Food Portion From Monocular Images \\ With Cross-Domain Feature Adaptation}
%
\name{Zeman Shao$^{*}$, Shaobo Fang$^{*}$, Runyu Mao$^{*}$, Jiangpeng He$^{*}$, Janine Wright$^{\dagger}$}
\secondlinename{Deborah Kerr$^{\dagger}$, Carol Jo Boushey$^{\ddagger}$, Fengqing Zhu$^{*}$}
\address{
$^{\star}$School of Electrical and Computer Engineering, Purdue University, West Lafayette, Indiana, USA \\
$^{\dagger}$School of Public Health, Curtin University, Perth, Western Australia \\
$^{\ddagger}$Cancer Epidemiology Program, University of Hawaii Cancer Center, Honolulu, Hawaii, USA
}

\maketitle

\begin{abstract}
We aim to estimate food portion size, a property that is strongly related to the presence of food object in 3D space, from single monocular images under real life setting.
Specifically, we are interested in end-to-end estimation of food portion size, which has great potential in the field of personal health management.
Unlike image segmentation or object recognition where annotation can be obtained through large scale crowd sourcing, it is much more challenging to collect datasets for portion size estimation since human cannot accurately estimate the size of an object in an arbitrary 2D image without expert knowledge.
To address such challenge, we introduce a real life food image dataset 
collected from a nutrition study where the groundtruth food energy (calorie) is provided by registered dietitians, and will be made available to the research community.
We propose a deep regression process for portion size estimation by combining features estimated from both RGB and learned energy distribution domains.
Our estimates of food energy achieved state-of-the-art with a MAPE of $11.47\%$, significantly outperforms non-expert human estimates by $27.56\%$.
\end{abstract}
%
\section{Introduction}
Measuring accurate dietary intake is challenging due to the high complexity of diet and the lack of unbiased and accurate tools. Harnessing the capabilities of mobile and vision-based technologies, new opportunity arise to improve the accuracy of dietary intake by capturing images of foods consumed at an eating occasion. However, estimating food portion size from a single-view image is an ill-posed problem. Image-based food portion estimation is not well defined and lacks appropriate datasets, which hinders further progress in this important topic area. 
In this paper, we propose to learn an object's portion size which is defined as the numeric value that is directly related to the spatial quantity of the object in world coordinates, such as an object's volume and weight. This information can then be used to calculate energy in kilocalories (kCal).
Specifically, we are interested in end-to-end solution for food portion size (in calories) from a single food image. 

We focus on the development of deep regression models for estimating food portion from an single monocular eating scene image.
The proposed method is evaluated on real life eating scene images collected from a dietary study.  
Estimating portion size from a single monocular image is an ill-posed problem since the spatial quantity is strongly correlated to a scene's 3D structure which is lost in the 2D image.
We chose such input format as majority of the data in real-world scenarios, especially those collected from mobile cameras, are single monocular RGB images. 
This is particularly true for applications such as image-based dietary assessment where visual data is collected mostly from mobile cameras and wearable sensors. 
In addition, to date estimation of energy consumed during a meal is undertaken by participants in nutrition studies using traditional recall and interview methods, which are time consuming and often lead to under reporting of energy intake \cite{schap2011}. More importantly, they are not suitable for everyday monitoring \cite{Thompson2017}. 

High quality datasets require many efforts from the research community and are the backbones for training-based techniques, e.g., ImageNet~\cite{ilsvrc_15} for object detection, MS-COCO~\cite{lin2014microsoft} for segmentation.
It is feasible to collect datasets for detection and segmentation of objects through crowd sourcing. 
However, it is difficult to obtain accurate food portion from the crowd based on RGB images, unless these values are recorded during image collection. 
To address this issue, we introduce an eating scene image dataset with known food weights provided by registered dietitians.
We describe the collection of our dataset in Section~\ref{sec:dataset}.

Based on experimental results shown in Section~\ref{sec:results}, directly using the original RGB image as input to train a deep regression model to predict portion size is not a feasible solution. 
Instead of training the deep regression end-to-end, we supervise the training on energy distribution as described in Section~\ref{sec:method}.
We show that with supervision on energy distribution, the deep regression model can be significantly improved compared to directly using original RGB image as input.
The model is end-to-end in inference since input is the original RGB image and output is the predicted numeric values of portion size.
In addition to the predicted energy distribution, we added features extracted from the original RGB input as detailed in Section~\ref{sec:intermediate_supervision}. 
This removes the sole dependency on predicted feature maps and adds robustness to proposed method by minimizing the impact of errors propagated from previous steps.
However, separate models trained in different domains have different feature space distributions, which leads to imbalanced feature weights in the concatenated features.
To adapt features extracted from different domains and to remove imbalance in feature space for joint regression, we extensively studied the use of normalization techniques~\cite{ulyanov2016instance, ba2016layer, wu2018group} in Section~\ref{sec:feature_adaptation}.
Our method showed significant improvement compared to results reported in prior art~\cite{fang2019end} for food energy estimation from monocular images. 
We show that our proposed method of learning food energy from monocular eating scene image with cross-domain feature adaptation achieved mean absolute percentage error (MAPE) of $11.47\%$ for energy estimation, significantly outperforms human estimation of $39.03\%$ MAPE.
We envision this would open doors to many possible applications for personal health management, and hold promise to approving accuracy while reducing participant and researcher burden.

\section{Related Work}


Food is an essential component of everyday life. 
The amount of food (\textit{i.e.}, food portion size) a person eats can directly impact his/her health.
%
Automatic estimation of food portion size from a monocular image is an open research problem and there exists several different approaches.
In~\cite{aizawa_2013}, food portion is divided into discrete serving sizes and food portion estimation is treated as a classification problem to determine a fixed serving size.
In~\cite{fang_2015}, pre-defined 3D food models are projected onto the scene to find the best fit with camera calibration.
In~\cite{yanai_2017}, a multi-task CNN is proposed to predict food class, ingredients, cooking instruction and food energy.
The dataset used in~\cite{yanai_2017} for food energy is obtained by web crawler from a cooking website and cannot be verified for its accuracy.
In addition, only one neuron in the last fully-connected layer is used for energy estimation, making it difficult to analyze the cause of error.
In~\cite{murphy_2015}, food volume is estimated from the predicted depth map of the eating scene.
The depth map is then converted to voxel representation which is used to estimated food volumes. 
Recently, an end-to-end approach for food energy estimation is proposed in~\cite{fang2019end}, where the concept of ``energy distribution map"~\cite{icip2018} replaces the ``depth map" in~\cite{murphy_2015} and the final food energy estimation is reported.
However, the estimation depends solely on the predicted ``energy distribution map", therefore any error from the ``energy distribution map" will propagate into the final portion size.
\section{Dataset Collection}
\label{sec:dataset}
The availability of public image datasets with annotation has resulted impressive success of deep methods for many computer vision tasks, such as object recognition and segmentation. 
To our best knowledge, there is no publicly available image dataset suitable for estimating food portion size since such groundtruth is not commonly available.
In order to address the lack of proper datasets, we introduce an eating scene image to food energy dataset, which contains eating scene images collected from a nutrition study, and the groundtruth energy per food item in each image is provided by registered dietitians.

\begin{figure}[t]
\subfigure[{Eating scene image $\mathbf{x}$.}]
{
\centering{\epsfig{figure=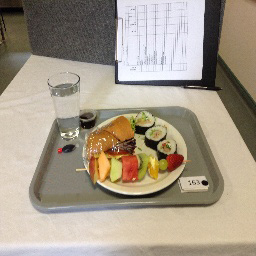, width = 3.9cm, height = 2.9cm}}
\label{fig:buffet_rgb}
}
\hfill
\subfigure[{Energy distribution map $\mathbf{y}$}]
{
\centering{\epsfig{figure=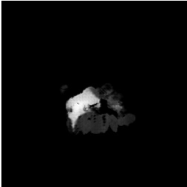, width = 3.9cm, height = 2.9cm}}
\label{fig:buffet_dis}
}
\vspace*{-0.5cm}
\caption{\textbf{An example eating scene data.} (a) RGB image, (b) corresponding energy distribution map. The associated groundtruth food energy is 606 kCal.
}
\vspace*{-0.5cm}
\label{fig:buffet_data}
\end{figure}

The eating scene images are collected as part of an image-assisted 24-hour dietary recall (24HR) study~\cite{boushey_2017new} conducted by registered dietitians.
The study participants are healthy volunteers aged between 18 and 70 years. A mobile app is used to capture images of the eating scenes for 3 meals (breakfast, lunch and dinner) over a 24-hour period.
Foods are provided in buffet style in which pre-weighted foods and beverages are served to the participants and the leftover foods and beverages are returned and weighted.
Participants are asked to capture both the before and after eating images for each meal.
Based on the known foods and their weight, food energy is calculated and used as groundtruth for evaluating the proposed method.
A complete pair of data include an RGB image and groundtruth food energy, as shown in Figure~\ref{fig:buffet_data}. 
This dataset contains 96 eating scene images, the groundtruth food energy ranges from 19.83 to 2,204.35 kilocalories (kCal), and the  mean is 717.52 kCal.

\begin{figure*}[ht]
\centering{\includegraphics[width=16.5cm]{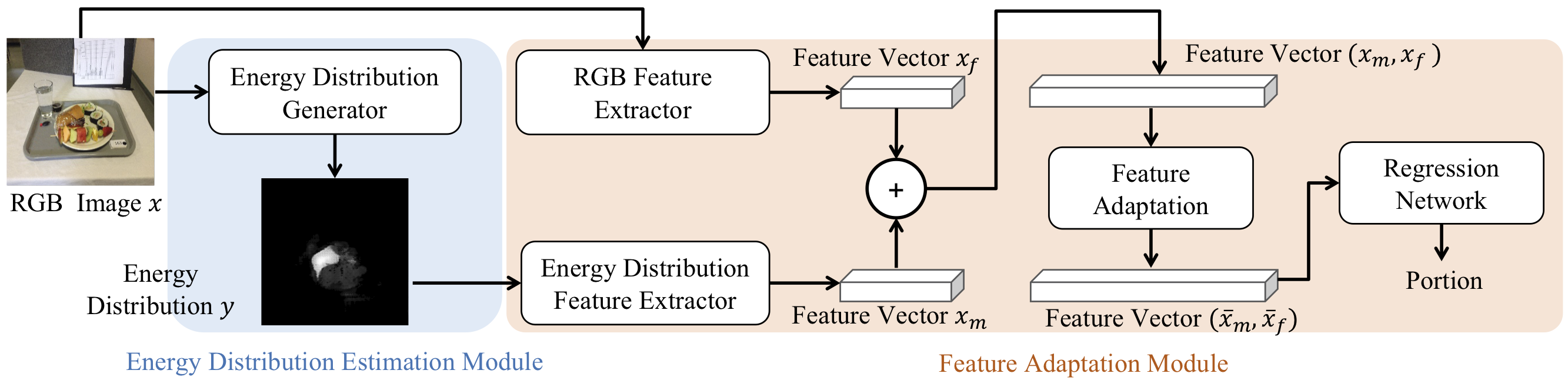}}
\caption{
\textbf{An overview of proposed method for estimating food portion from monocular images.} It consists of two modules: (1) generate intermediate result of energy distribution map, (2) combine and normalize features extracted from RGB and energy distribution domains for efficient regression. (Best viewed in color)}
\vspace*{-0.5cm}
\label{fig:vol_sys}
\end{figure*}

\section{Learning Object's Spatial Quantity With Cross-Domain Feature Adaptation}
\label{sec:method}
Since portion size is directly related to the spatial quantity of objects in 3D space, it can be viewed as a regression task which aims to estimate numeric values based on the input image.
Directly using the original RGB image as input to train a deep regression model to predict portion size (a single numeric value) is not feasible. 
For example, if the input image is of size $256 \times 256 \times 3$, direct approach would map $\mathcal{R}^{256\times256\times3} \to \mathcal{R}^{1\times1\times1}$ and it is difficult to learn such a mapping.
To overcome this challenge, we propose a deep regression process that utilizes adapted features from the input RGB image and predicted energy distribution map (Figure~\ref{fig:buffet_dis}), illustrated in Figure~\ref{fig:vol_sys}. 
Our method consists of two modules: 
    \textbf{Energy Distribution Estimation}: goal is to generate an accurate energy distribution, based on the input RGB image.
    \textbf{Feature Adaptation}: goal is to combine the features extracted from RGB and energy distribution domains and normalize them for efficient regression.

\subsection{Intermediate Result - Energy Distribution}
\label{sec:intermediate_supervision}
Our goal is to obtain an end-to-end regression model where the input is an image, and the output is a numeric value related to spatial quantity, \textit{i.e.}, food energy.
The end-to-end is only required for the inference stage.
In the training stage, the model does not need to be one neural network trained simultaneously.
It has been shown in~\cite{intermediate_concept} that training on intermediate concept can improve the performance of trained models.
Kim \textit{et al.} \cite{yuvill_2019} showed that models trained on synthetic data could b1e transferred to real world videos for activity recognition.  
Myers \textit{et al.} \cite{murphy_2015} showed that object's volume can be estimated using voxel representation from predicted depth images.
Summing each voxel volume would then provide the volume of the objects in the scene.

Since there is no mapping available from a RGB image to the spatially distribution of food energy in the scene,  a synthetic intermediate result of ``energy distribution" image was proposed in \cite{icip2018}. An ``energy distribution" image has pixel-to-pixel correspondences and the weights at different pixel locations represent how food energy is distributed in the eating scene. 
For example, pixels corresponding to steak have much higher weight than pixels of apple. 
In the subsequent work~\cite{fang2019end}, an end-to-end system of estimating eating scene food energy is proposed, where the input is the eating scene RGB image.
We follow the architecture proposed in~\cite{fang2019end} and train a ``energy distribution" estimation module using a Generative Adversarial Networks in conditional settings~\cite{pix2pix}.
The regression model is then trained using predicted ``energy distribution image" as input as shown in Figure ~\ref{fig:vol_sys}.
The portion estimation in the inference stage is end-to-end where input is the original eating scene RGB image, and output is the total food energy estimated from the scene.

The mapping from original RGB space to the energy distribution space as described in Figure~\ref{fig:vol_sys} is analogous to performing dimension reduction in feature spaces.
Without additional features from original RGB images, the regression model depends solely on the predicted mappings generated, and errors from the estimated energy distribution propagating into the subsequent regression module as discussed in \cite{fang2019end}.

\subsection{Features Adaptations from Multiple Domains}
\label{sec:feature_adaptation}
To address this sole dependency on prior energy distribution, which may create a bottleneck for estimation accuracy, we propose to combine features from both the predicted energy distribution and the original RGB images to enables more robust estimation with additional features.
We use the features extracted from a modified ResNet-50~\cite{resnet} pre-trained on ImageNet~\cite{ilsvrc_15} where the last fully connected layer is removed as the additional features for volume estimation.
Similarly, we use the features extracted from this modified ResNet-50 trained on Recipe1M~\cite{salvador2017learning} as the additional features for food energy estimation.
The features from Recipe1M~\cite{salvador2017learning} was originally used for the joint learning of ingredients and cooking instructions which achieved state-of-the-art results. Therefore, they are good representations of food image features and are suitable for transfer learning between two tasks both are related to food images. 

We denote the features extracted from energy distribution as $\textbf{x}_{m}$ and the features extracted from original RGB image as $\textbf{x}_f$.
The $\textbf{x}_{m}$ and $\textbf{x}_{f}$ are extracted from separate models trained in different domains, thus have significant differences reflected by their mean and variance.
Simply concatenating the features $(\mathbf{x}_{m}, \mathbf{x}_{f})$ (of dimension $R^{C \times 1}$) and applying fully-connected layers causes performance degradation of the regression model. 
To adapt the features extracted from different domains and to remove imbalance in feature space for joint regression, we extensively studied the use of normalization techniques.
We used z-score as the baseline normalization such that:
\begin{equation}
    \mathbf{\bar{x}}_{f} = \sigma_{\mathbf{x}_{m}} \cdot  \frac{\mathbf{x}_{f} - \mu_{\mathbf{x}_{f}}}{\sigma_{\mathbf{x}_{f}}}  + \mu_{\mathbf{x}_m}
    \label{eq:z-score}
\end{equation}
where $\mu_{\mathbf{x}_{f}}$ and $\sigma_{\mathbf{x}_{f}}$ in Equation~\ref{eq:z-score} are the mean and standard deviation of $\textbf{x}_{f}$, $\mu_{\mathbf{x}_{m}}$ and $\sigma_{\mathbf{x}_{m}}$ are for $\mathbf{x}_m$ accordingly. 
The concatenated features after z-score normalization, $(\mathbf{x}_{m}, \mathbf{\bar{x}}_{f})$, are now used for the subsequent regression model.
Since z-score normalization parameters are not learnable, the capacity of the technique is limited.

Common normalization techniques that have learnable parameters include Batch Normalization (BN)~\cite{ioffe2015batch}, Instance Normalization (IN)~\cite{ulyanov2016instance}, Layer Normalization (LN)~\cite{ba2016layer} and Group Normalization (GN)~\cite{wu2018group}.
Since small mini-batches offer a regularizing effect~\cite{wilson2003general}, we use a mini-batch size of 2 in training. As shown in~\cite{wu2018group}, BN does not work well with small mini-batch sizes.
As concatenated features $(\mathbf{x}_{m}, \mathbf{x}_{f})$ have dimension of $R^{C \times 1}$, IN is not suitable since estimating the mean and variance for a single value is not meaningful.
Both LN and GN that are independent of mini-batch size overcome the drawback of BN that exploits the batch dimension.
Since the feature $(\mathbf{x}_{m}, \mathbf{x}_{f})$ is concatenated from two different domains, the number of groups for GN used in this work is pre-defined as 2. 
LN and GN are defined as:
\begin{equation}
    y_i = \gamma \hat{x}_i + \beta, \text{\ where \ } \hat{x}_i = \frac{1}{\sigma_i} (x_i - \mu_i) 
\end{equation}
where $\gamma$ and $\beta$ are learnable parameters, $\hat{x}_i$ is the normalized source domain sample for $x_i$ and $y_i$ is the mapped sample based on learned normalization.
More specifically, since the concatenated feature vector $(\mathbf{x}_{m}, \mathbf{\bar{x}}_{f})$ has dimension $R^{C \times 1}$, we define $i = (i_N , i_C)$, a 2D vector indexing the features in $(N, C)$, where $N$ is the batch axis, $C$ is the channel axis.
$\sigma_i$ and $\mu_i$ are defined as:
\begin{equation}
    \mu_i = \frac{1}{m} \sum_{k \in \mathcal{S}_i} x_k, \ \ \ \sigma_i = \sqrt{\frac{1}{m} \sum_{k \in \mathcal{S}_i} (x_k - \mu_i)^2 + \epsilon}
\label{eq:define_mu_std}    
\end{equation}
LN computes $\mu_i$ and $\sigma_i$ across $(C)$ channels where $S_i =\{ k \ | \ k_N = i_N \}$ and GN computes $\mu$
and $\sigma$ in a set $\mathcal{S}_i$ that is defined as $\mathcal{S}_i = \{ k \ | \  k_N = i_N, \lfloor{\frac{k_C}{C/G}} \rfloor = \lfloor{\frac{i_C}{C/G}} \rfloor  \} $, where $G$ is the number of groups, and $\epsilon$ is a small constant.
\begin{figure}[t]
\centering
\includegraphics[width = 5.3cm]{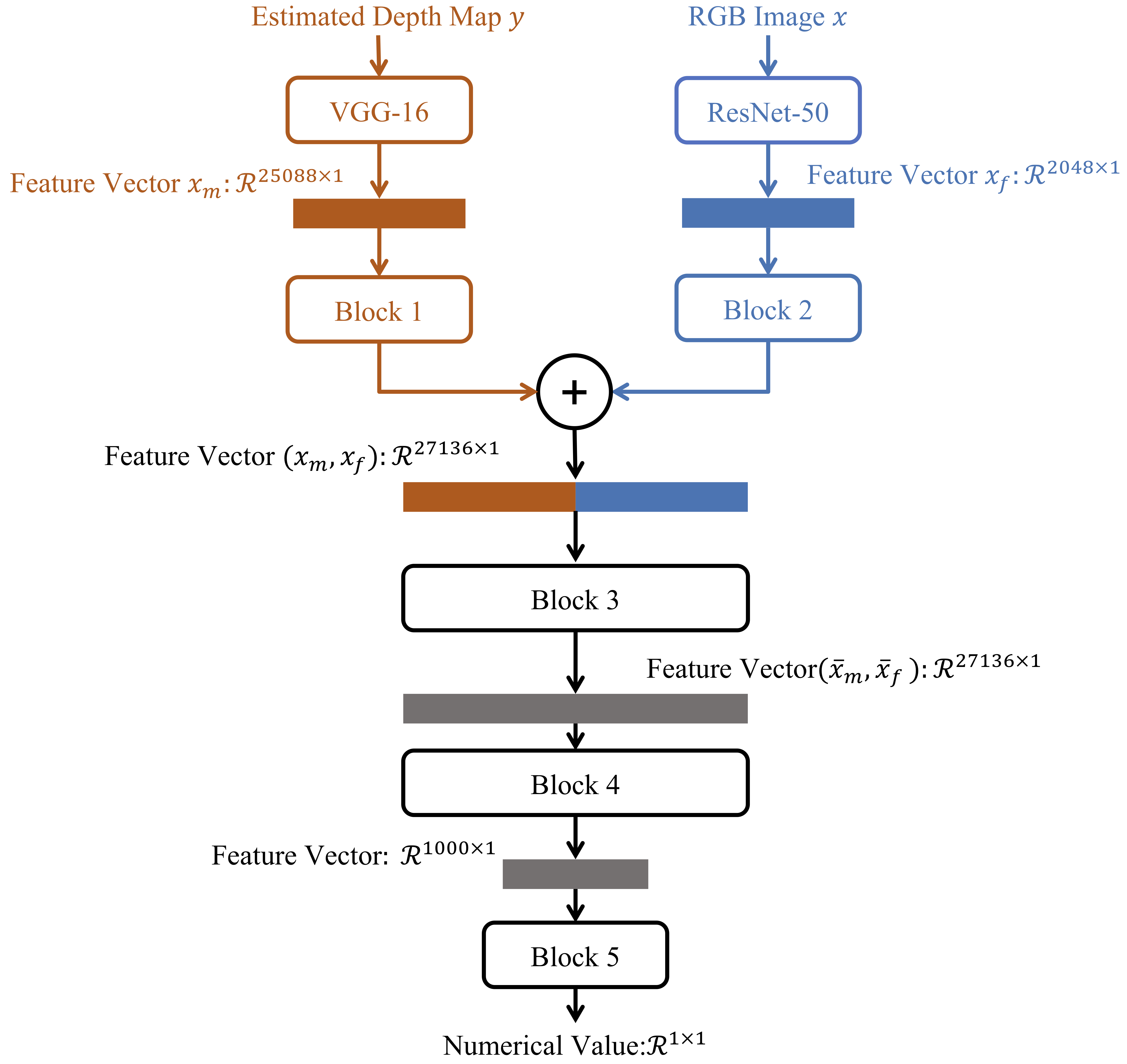}
\caption{\textbf{The architecture for feature adaptation module using LN and LN+GN.} Each numbered block represents a group of few stacked layers. (Best viewed in color)}
\vspace*{-0.5cm}
\label{fig:adaptation}
\end{figure}
The architecture for feature adaptation module is illustrated in Figure~\ref{fig:adaptation}.
We investigated three normalization methods, namely z-score normalization, LN only and LN+GN\footnote{Implementation details of feature adaptation modules (z-score, LN, LN+GN) are described in the supplementary material.}.
We denote the normalized concatenated feature with learnable parameters as: $(\mathbf{\bar{x}_m}, \mathbf{\bar{x}_f})$.
Details for the feature adaptation modules (z-score, LN, LN+GN) are described below:
\begin{itemize}
  \item  \textbf{Normalization using z-score}: 
        \textbf{Block 3}: we normalize the concatenated features $(\mathbf{x}_{m}, \mathbf{x}_{f})$ using z-socre normalization as described in Equation~\ref{eq:z-score}.
        \textbf{Block 4}: FC-1000 + ReLU + Dropout.
        \textbf{Block 5}: FC-1.
    \item \textbf{Normalization using LN and LN+GN}: 
      \textbf{Block 1}: LN Layer + ReLU + Dropout.
      \textbf{Block 2}: LN Layer + ReLU + Dropout.
      \textbf{Block 5}: FC-1.
        \item   \textbf{For LN}: 
        \textbf{Block 3}: LN Layer + ReLU + Dropout.
        \textbf{Block 4}: FC-1000 + LN Layer + ReLU + Dropout.
        \item  \textbf{For LN+GN}: 
      \textbf{Block 3}: GN Layer + ReLU + Dropout.
      \textbf{Block 4}: FC-1000 + GN Layer + ReLU + Dropout.
\end{itemize}
We set $\epsilon=1e-5$ for LN Layer, $\epsilon=1e-5$ and $G=2$ for GN Layer, and Dropout with probability $p=0.5$.
\section{Experimental Results}
\label{sec:results}
\begin{figure*}[t]
\subfigure[{}]
{
\label{fig:vol_rgb}
\centering{\epsfig{figure=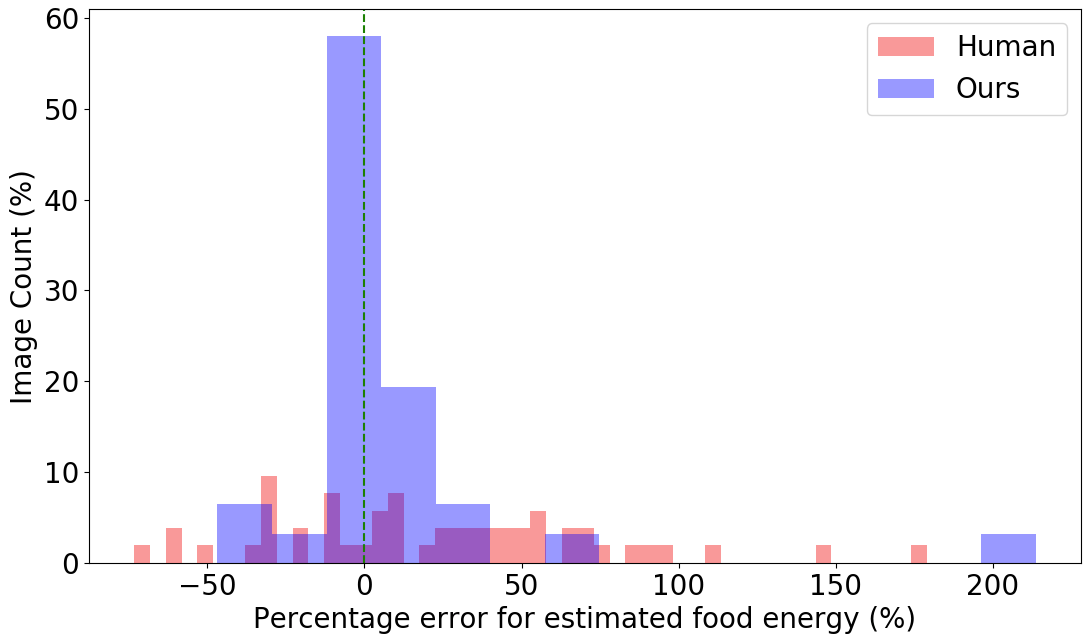, height=3.3cm}}
}
\hfill
\subfigure[{}]
{
 \centering{\epsfig{figure=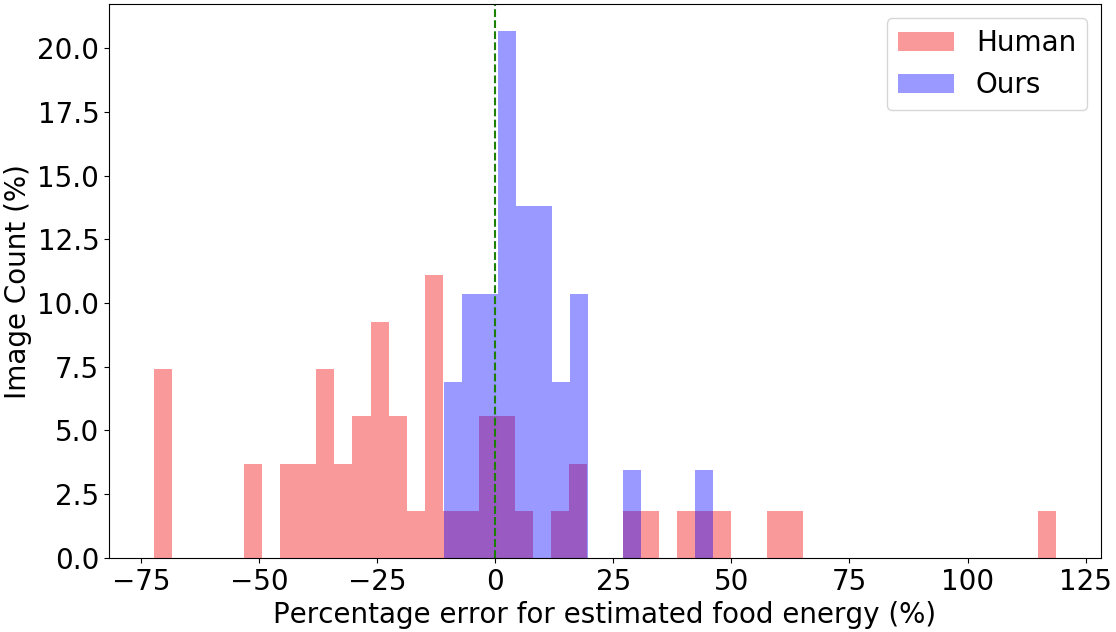, height = 3.3cm}}
\label{fig:vol_depth}
}
\hfill
\subfigure[{}]
{
 \centering{\epsfig{figure=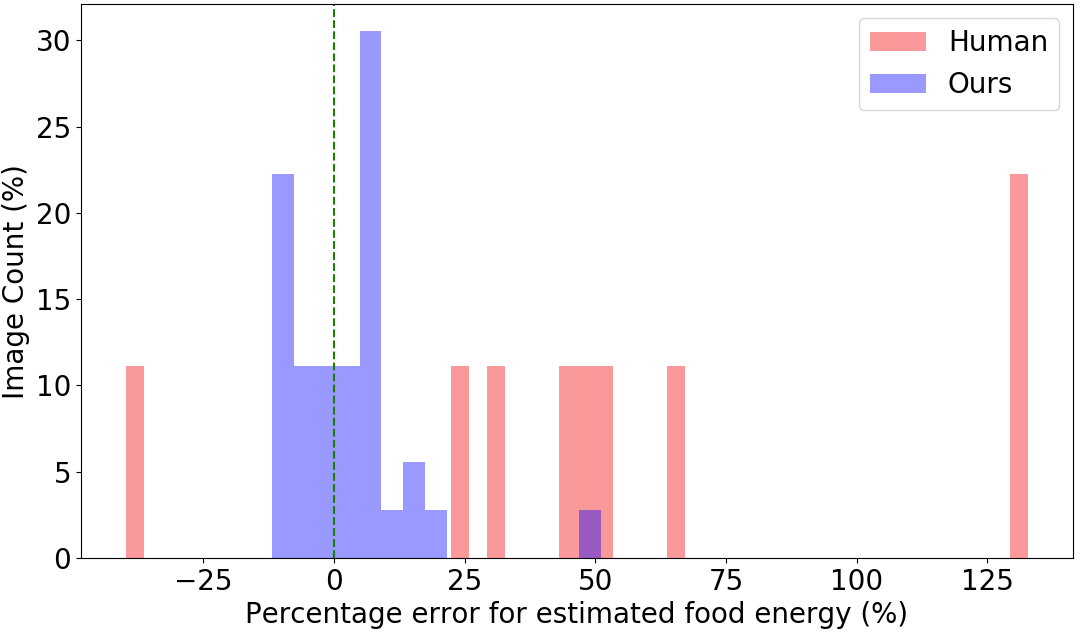, height = 3.3cm}}
\label{fig:vol_depth}
}
\vspace*{-0.5cm}
\caption{Percentage error distribution of (a) breakfast, (b) lunch, and (c) dinner meals estimated by human and proposed method. (Best viewed in color)
}
\vspace*{-0.5cm}
\label{fig:human_hist}
\end{figure*} 
In this section, we evaluate the proposed method using the dataset introduced in Section~\ref{sec:dataset}, \textit{i.e.}, the eating scene image to food energy dataset.
The eating scene images are collected as part of an image-assisted 24-hour dietary recall (24HR) study~\cite{boushey_2017new} conducted by registered dietitians which required a structured interview of the participants.
Therefore, the number of images collected for this dataset is limited. We applied data augmentation techniques, such as rotating, cropping, and flipping, to the training and testing sets to expand the dataset.
In the end, we have 864 eating scene images for training and 96 images for testing.

We use two common regression metrics, mean absolute error (MAE) and mean absolute percentage error (MAPE), defined as,
\begin{equation}
\text{MAE} = \frac{1}{N}\sum^{N}_{i=1} |\tilde{w}_i - \bar{w}_i|
\end{equation}
\begin{equation}
\text{MAPE} = \frac{100\%}{N}\sum^{N}_{i=1} \frac{|\tilde{w}_i - \bar{w}_i|}{\bar{w}_i}
\end{equation}
where $\tilde{w}_i$ is the estimated portion size of the $i$-th image, $\bar{w}_i$ is the groundtruth portion size of $i$-th image and $N$ is the number of testing images. The average MAE and MAPE reported below are calculated based on the models obtained from the last 50 training epochs.
\begin{table}[t]
\fontsize{9}{11}\selectfont
\begin{center}
\caption{Experimental results for food energy estimation on eating scene dataset. $\mathbf{x}_f$: RGB feature; $\mathbf{x}_m$: Energy distribution feature; $(\mathbf{x}_m, \mathbf{x}_f)$: Concatenated features; $(\mathbf{x}_m, \mathbf{\bar{x}}_f)$ and $(\mathbf{\bar{x}}_m, \mathbf{\bar{x}}_f)$: Concatenated features after normalization}
\vspace{0.2cm}
\label{tab:cal_eval}
\begin{tabular}{>{\centering}p{2.5cm} >{\centering}p{1.6cm} c}
\hline
Method                                     & MAE (kCal)      & MAPE (\%)     \\
\hline
$\mathbf{x}_f$                               & 292.35 & 151.33 \\
$\mathbf{x}_m$                              & 77.76  & 17.63 \\
$(\mathbf{x}_m, \mathbf{x}_f)$              & 110.84 & 99.36 \\ 
\hline
$(\mathbf{x}_m, \mathbf{\bar{x}}_f)$: z-score             & 75.15 & 22.24 \\
$(\mathbf{\bar{x}}_m, \mathbf{\bar{x}}_f)$: LN+GN         & 57.75 & 16.90 \\
$(\mathbf{\bar{x}}_m, \mathbf{\bar{x}}_f)$: LN            & \textbf{56.22} & \textbf{11.47} \\
\hline 
\end{tabular}
\end{center}
\vspace{-0.5cm}
\end{table}
We use the method proposed in~\cite{fang2019end} as the baseline and evaluate the proposed method on the eating scene image to food energy dataset by conducting multiple trials and report the average performance.
We compared results of using 3 different normalization techniques, z-score, and LN and LN+GN which are summarized in Table~\ref{tab:cal_eval}.
The concatenated feature normalized using z-score $(\mathbf{x}_m, \mathbf{\bar{x}}_f)$ has slightly improved results by MAE over the baseline which use only $\mathbf{x}_m$, predicted energy distribution. 
We further studied the use of normalization techniques with learnable parameters and focused on Layer Norm and Group Norm. 
We denote the concatenated features normalized by LN/LN+GN as $(\mathbf{\bar{x}_m}, \mathbf{\bar{x}_f})$.
We showed that LN achieved the best performance as indicated by lowest MAE and MAPE, $56.22$ kCal and $11.47\%$ respectively, compared to other methods. 
Addition analysis of the results are discussed in Section~\ref{ssec:ablation}.

\subsection{Ablative Analysis}
\label{ssec:ablation}
To investigate the influence of each module, we performed ablative analysis by remove them one at a time and evaluated the performance using the two datasets.
We summarize the results for food energy estimation in Table~\ref{tab:cal_eval}.
We use results from \cite{fang2019end} as the baseline method.
We achieved a MAE of 77.76 kCal and MAPE of 17.63\% using the baseline and it is much better than directly using RGB features as shown in Table \ref{tab:cal_eval}, where the MAE and MAPE are 292.35 kCal and 151.33\%, respectively. 
As shown in Table~\ref{tab:cal_eval}, directly using concatenating features of $(\mathbf{x}_m, \mathbf{x}_f)$ causes performance degradation both in MAE and MAPE, 110.84 kCal and 99.36\% respectively, as the features from two domains have significant differences reflected by their mean and variance.

\subsection{Comparison to human estimates}
Given the high accuracy of portion size estimation for eating scene food energy, we are interested in how well our automatic technique compares to human performance in which the participants of this nutrition study estimated the portion size using image-assisted 24-hr dietary recall. 
We compared our food energy estimates to participants' estimates from the same nutrition study .
The study participants used a mobile app to capture images of the eating scene for 3 meals (breakfast, lunch and dinner) over a 24-hour period.
At the end of the day, participants estimated the portion size of the meal they consumed in a structured interview while viewing the captured images~\cite{boushey_2017new}, we calculated MAPE and MAE of participant estimates based on the groundtruth food energy they consumed.

The MAPE of participant estimates is 39.03\%, and the MAE is 286.37 kCal.
We compared participant estimates to our best result which is predicted by concatenating $\mathbf{x}_f$ and $\mathbf{x}_m$ followed by LN which achieved MAE of 56.22 kCal and MAPE of 11.47\%.  This is also illustrated in Figure~\ref{fig:human_hist} for breakfast, lunch and dinner meals separately. We observe that our proposed method outperforms participant estimates which indicates that estimating portion size accurately from a single monocular image is a challenging task for human. 
Note also that participants were not able to recall all items they consumed, particularly for sauces and salad dressings.

\section{Conclusion}
In this work, we proposed an end-to-end framework for learning food portion from monocular images which is validated on a new real life eating scene image to food energy dataset with groundtruth portion size in food energy.
We showed that with supervision on energy distribution map, a deep regression model can be significantly improved compared to directly using original RGB image as input.
We extensively investigated different normalization techniques when adapting features from different domains, and showed that normalization can further improve the estimation accuracy. 
Our method achieved state-of-the-art accuracy for the challenging real life food image dataset with a MAE of $56.22$ calories and a MAPE of $11.47\%$, surpassing human estimates of $39.03\%$ MAPE. 
Our preliminary results showed promising applications for automated dietary and health monitoring.
\bibliographystyle{IEEEbib}
\bibliography{main}

\begin{thebibliography}{10}

\bibitem{schap2011}
T.~E. Schap, B.~L. Six, E.~J. Delp, D.~S. Ebert, D.~A. Kerr, and C.~J. Boushey,
\newblock ``Adolescents in the united states can identify familiar foods at the
  time of consumption and when prompted with an image 14 h postprandial, but
  poorly estimate portions,''
\newblock {\em Public Health Nutrition}, vol. 1, no. 1, pp. 1--8, July 2011.

\bibitem{Thompson2017}
F.~E. Thompson and A.~F. Subar,
\newblock ``{Dietary Assessment Methodology},''
\newblock in {\em Nutrition in the Prevention and Treatment of Disease}, pp.
  5--48. Elsevier, 2017.

\bibitem{ilsvrc_15}
O.~Russakovsky, J.~Deng, H.~Su, J.~Krause, S.~Satheesh, S.~Ma, Z.~Huang,
  A.~Karpathy, A.~Khosla, M.~Bernstein, A.~Berg, and F.~Li,
\newblock ``Imagenet large scale visual recognition challenge,''
\newblock {\em International Journal of Computer Vision}, vol. 115, no. 3, pp.
  211 -- 252, 2015.

\bibitem{lin2014microsoft}
T.~Lin, M.~Maire, S.~Belongie, J.~Hays, P.~Perona, D.~Ramanan, P.~Doll{\'a}r,
  and C.~L. Zitnick,
\newblock ``Microsoft coco: Common objects in context,''
\newblock pp. 740--755, 2014.

\bibitem{ulyanov2016instance}
D.~Ulyanov, A.~Vedaldi, and V.~Lempitsky,
\newblock ``Instance normalization: The missing ingredient for fast
  stylization,''
\newblock 2016.

\bibitem{ba2016layer}
J.~L. Ba, J.~R. Kiros, and G.~E. Hinton,
\newblock ``Layer normalization,''
\newblock 2016.

\bibitem{wu2018group}
K.~He Y.~Wu,
\newblock ``Group normalization,''
\newblock {\em arXiv preprint arXiv:1803.08494}, 2018.

\bibitem{fang2019end}
S.~Fang, Z.~Shao, D.~A. Kerr, C.~J. Boushey, and F.~Zhu,
\newblock ``An end-to-end image-based automatic food energy estimation
  technique based on learned energy distribution images: Protocol and
  methodology,''
\newblock {\em Nutrients}, vol. 11, no. 4, pp. 877, 2019.

\bibitem{aizawa_2013}
K.~Aizawa, Y.~Maruyama, H.~Li, and C.~Morikawa,
\newblock ``Food balance estimation by using personal dietary tendencies in a
  multimedia {Food Log},''
\newblock {\em IEEE Transactions on Multimedia}, vol. 15, no. 8, pp. 2176 --
  2185, December 2013.

\bibitem{fang_2015}
S.~Fang, C.~Liu, F.~Zhu, E.~Delp, and C.~Boushey,
\newblock ``{Single-}view food portion estimation based on geometric models,''
\newblock {\em Proceedings of the IEEE International Symposium on Multimedia},
  pp. 385--390, December 2015,
\newblock {Miami, FL}.

\bibitem{yanai_2017}
T.~Ege and K.~Yanai,
\newblock ``Image-based food calorie estimation using knowledge on food
  categories, ingredients and cooking directions,''
\newblock {\em Proceedings of the Workshops of ACM Multimedia on Thematic}, pp.
  367--375, 2017,
\newblock {Mountain View, CA}.

\bibitem{murphy_2015}
A.~Myers, N.~Johnston, V.~Rathod, A.~Korattikara, A.~Gorban, N.~Silberman,
  S.~Guadarrama, G.~Papandreou, J.~Huang, and K.~Murphy,
\newblock ``{Im2Calories:} towards an automated mobile vision food diary,''
\newblock {\em Proceedings of the IEEE International Conference on Computer
  Vision}, December 2015,
\newblock {Santiago, Chile}.

\bibitem{icip2018}
S.~Fang, Z.~Shao, R.~Mao, C.~Fu, E.~J. Delp, F.~Zhu, D.~A. Kerr, and C.~J.
  Boushey,
\newblock ``Single-view food portion estimation: learning image-to-energy
  mappings using generative adversarial networks,''
\newblock {\em Proceedings of the IEEE International Conference on Image
  Processing}, pp. 251--255, October 2018,
\newblock Athens, Greece.

\bibitem{boushey_2017new}
C.~J. Boushey, M.~Spoden, F.~M. Zhu, E.~J. Delp, and D.~A. Kerr,
\newblock ``New mobile methods for dietary assessment: review of image-assisted
  and image-based dietary assessment methods,''
\newblock {\em Proceedings of the Nutrition Society}, vol. 76, no. 3, pp.
  283--294, August 2017.

\bibitem{intermediate_concept}
C.~{Li}, M.~Z. {Zia}, Q.~{Tran}, X.~{Yu}, G.~D. {Hager}, and M.~{Chandraker},
\newblock ``Deep supervision with intermediate concepts,''
\newblock {\em IEEE Transactions on Pattern Analysis and Machine Intelligence},
  vol. 41, no. 8, pp. 1828--1843, Aug 2019.

\bibitem{yuvill_2019}
T.~S. {Kim}, M.~{Peven}, W.~{Qiu}, A.~{Yuille}, and G.~D. {Hager},
\newblock ``Synthesizing attributes with unreal engine for fine-grained
  activity analysis,''
\newblock {\em 2019 IEEE Winter Applications of Computer Vision Workshops}, pp.
  35--37, Jan 2019.

\bibitem{pix2pix}
P.~Isola, J.~Y. Zhu, T.~Zhou, and A.~A. Efros,
\newblock ``Image-to-image translation with conditional adversarial networks,''
\newblock {\em Proceedings of the IEEE Conference on Computer Vision and
  Pattern Recognition}, pp. 5967--5976, July 2017,
\newblock {Honolulu, HI}.

\bibitem{resnet}
K.~He, X.~Zhang, S.~Ren, and J.~Sun,
\newblock ``Deep residual learning for image recognition,''
\newblock {\em Proceedisng of the IEEE Conference on Computer Vision and
  Pattern Recognition}, pp. 770--778, June 2016,
\newblock {Las Vegas, NV}.

\bibitem{salvador2017learning}
A.~Salvador, N.~Hynes, Y.~Aytar, J.~Marin, F.~Ofli, I.~Weber, and A.~Torralba,
\newblock ``Learning cross-modal embeddings for cooking recipes and food
  images,''
\newblock {\em Proceedings of the IEEE Conference on Computer Vision and
  Pattern Recognition}, July 2017,
\newblock {Honolulu, HI}.

\bibitem{ioffe2015batch}
S.~Ioffe and C.~Szegedy,
\newblock ``Batch normalization: Accelerating deep network training by reducing
  internal covariate shift,''
\newblock 2015.

\bibitem{wilson2003general}
R.~Wilson and T.~Martinez,
\newblock ``The general inefficiency of batch training for gradient descent
  learning,''
\newblock {\em Neural networks}, vol. 16, no. 10, pp. 1429--1451, 2003.

\end{thebibliography}

\end{document}


\sloppy

\def\x{{\mathbf x}}
\def\L{{\cal L}}

\title{\textit{Supplemental Material} \\ Towards Learning Food Portion From Monocular Images \\ With Cross-Domain Feature Adaptation \\
}
%
\name{Anonymous ICME submission}
\address{}

\maketitle

\section{Eating Scene Image to Food Energy Dataset}
This dataset includes RGB eating scene image and groundtruth food energy. 
The eating scene images are collected as part of an image-assisted 24-hour dietary recall (24HR) study conducted by registered dietitians.
Foods are provided in buffet style in which pre-weighted foods and beverages are served to the participants and the leftover foods and beverages are returned and weighted.
Based on the known foods and their weight, food energy is calculated and used as groundtruth.
The 96 eating scene images are located in the \textit{images} folder, the groundtruth food energy and human estimates are described in \textit{info.txt} file by following format:

\textless File Name\textgreater~\textless Groundtruth Energy (kCal)\textgreater